# Multi-Circle Detection on Images Using Artificial Bee Colony (ABC) Optimization

Erik Cuevas, Felipe Sención-Echauri, Daniel Zaldivar and Marco Pérez-Cisneros

Departamento de Ciencias Computacionales
Universidad de Guadalajara, CUCEI
Av. Revolución 1500, Guadalajara, Jal, México
{erik.cuevas, felipe.sencion, [1]daniel.zaldivar, marco.perez}@cucei.udg.mx

**Abstract**

Hough transform (HT) has been the most common method for circle detection, exhibiting robustness, but adversely demanding considerable computational effort and large memory requirements. Alternative approaches include heuristic methods that employ iterative optimization procedures for detecting multiple circles. Since only one circle can be marked at each optimization cycle, multiple executions must be enforced in order to achieve multi-detection. This paper presents an algorithm for automatic detection of multiple circular shapes that considers the overall process as a multi-modal optimization problem. The approach is based on the artificial bee colony (ABC) algorithm, a swarm optimization algorithm inspired by the intelligent foraging behavior of honey bees. Unlike the original ABC algorithm, the proposed approach presents the addition of a memory for discarded solutions. Such memory allows holding important information regarding other local optima which might have emerged during the optimization process. The detector uses a combination of three non-collinear edge points as parameters to determine circle candidates. A matching function (nectar- amount) determines if such circle candidates (bee-food-sources) are actually present in the image. Guided by the values of such matching functions, the set of encoded candidate circles are evolved through the ABC algorithm so that the best candidate (global optimum) can be fitted into an actual circle within the edge-only image. Then, an analysis of the incorporated memory is executed in order to identify potential local optima, i.e., other circles. The proposed method is able to detect single or multiple circles from a digital image through only one optimization pass. Simulation results over several synthetic and natural images, with a varying range of complexity, validate the efficiency of the proposed technique regarding its accuracy, speed, and robustness.

## 1. Introduction

The problem of detecting circular features holds paramount importance for image analysis in industrial applications such as automatic inspection of manufactured products and components, aided vectorization of drawings, target detection, etc. [1]. Solving common challenges for object localization is normally approached by two techniques: deterministic and stochastic. The former includes the application of Hough transform-based methods [2], the use of geometric hashing and other template or model-based matching techniques [3, 4]. On the other hand, stochastic techniques include random sample consensus techniques [5], simulated annealing [6] and genetic algorithms (GA) [7].

Template and model matching techniques were the first approaches to be applied to shape detection. Although several methods have now been developed for solving such a problem [8], shape coding techniques and a combination of shape properties have been successfully tested on representing different objects. Their main drawbacks are related to the contour extraction step from real images and to their deficiencies in dealing with pose invariance except for very simple objects.

The circle detection in digital images is commonly solved through the Circular Hough Transform [9]. A typical Hough-based approach employs an edge detector and some edge information to infer locations and radii values. Peak detection is then performed by averaging, filtering and histogramming within the transform space. Unfortunately, such an approach requires a large storage space as the 3-D cells include parameters ($x, y, r$) that augment the computational complexity and yield a low processing speed. The accuracy of parameters for the extracted circles is poor, particularly under noisy conditions [10].

In the particular case of a digital image holding a significant width and height, and some densely populated edge pixels, the required processing time for Circular Hough Transform makes it prohibitive to be deployed in real time applications. In order to overcome such a problem, some other researchers have proposed new approaches following Hough transform principles, yielding the probabilistic Hough transform [11], the randomized Hough transform (RHT) [12], the fuzzy Hough transform [13] and some other topics as is widely discussed by Becker in [14].

As an alternative to Hough Transform-based techniques, the problem of shape recognition in computer vision has also been handled through optimization methods. Ayala–Ramirez *et al.* presented a GA based circle detector [15] that is capable of detecting multiple circles over real images, but fails frequently while detecting imperfect circles. On the other hand, Dasgupta et al. [16] have recently proposed an automatic circle detector using the bacterial foraging optimization algorithm (BFOA) as optimization procedure. However, both methods employ an iterative scheme to achieve multiple-circle detection, which executes the algorithm as many times as the number of circles to be found demands. Only one circle can be found at each run yielding quite a long execution time.

An impressive growth in the field of biologically inspired meta-heuristics for search and optimization has emerged during the last decade. Some bio-inspired examples like genetic algorithm (GA) [17] and differential evolution (DE) [18] have been applied to solve complex optimization problems, while swarm intelligence (SI) has recently attracted interest from several fields. The SI core lies in the analysis of the collective behavior of relatively simple agents working on decentralized systems. Such systems typically gather an agent's population that can communicate with each other while sharing a common environment. Despite a non-centralized control algorithm regulating its behavior, the agent can solve complex tasks by analyzing a given global model and harvesting cooperation to other agents. Therefore, a novel global behavior evolves from interaction among agents as can be seen in case of ant colonies, animal herding, bird flocking, fish schooling, honey bees, bacteria, and many more. Swarm-based algorithms, such as particle swarm optimization [19], ant colony optimization [20] and bacterial foraging optimization algorithm (BFOA) [21] have already been successfully applied to several engineering applications.

Karaboga has recently presented one bee-swarm algorithm for solving numerical optimization problems, which is known as the artificial bee colony (ABC) method [22]. Inspired by the intelligent foraging behavior of a honeybee swarm, the ABC algorithm consists of three essential components: food source positions, nectar- amount and different honey bee classes. Each food source position represents a feasible solution for the problem under consideration, and the nectar-amount of a food source represents the quality of such solution corresponding to its fitness value. Each class of bees symbolizes one particular operation for generating new candidate food source positions (candidate solutions).

The ABC algorithm starts by producing a randomly distributed initial population (food source locations). After initialization, an objective function evaluates whether such candidates represent an acceptable solution (nectar-amount) or not. Guided by the values of such an objective function, the candidate solutions are evolved through different ABC operations (honey bee types). When the fitness function (nectar-amount) cannot be further improved after a maximum number of cycles, its related food source is assumed to be abandoned and replaced by a new randomly chosen food source location. However, in order to contribute towards the solution of multi-modal optimization problems, our proposal suggests that such abandoned solutions are not to be discarded; instead, they are to be arranged into a so-called "exhausted-source memory" that contains valuable information regarding global and local optima that have been emerging with the evolution of optimization.

Although ABC draws several similarities with other bio-inspired algorithms, there are some significant issues to be discussed: ABC does not depend upon the best member within the population in order to update the particle's motion as is done by PSO [23]; it does not require all particles for computing parameters such as the pheromone concentration, which determines the overall performance, as is demanded by ACO [24]. In contrast, ABC uses randomly chosen particles to calculate new motion vectors, contributing towards augmenting the population diversity. Similar to DE, ABC does require a selection operation that allows individuals to access a fair chance of being elected for recombination (diversity). However, ABC holds a second modification operation that follows a random "roulette selection", allowing some privileges for best located solutions and augmenting the convergence speed [25]. In contrast to the local particle modifications executed by BFOA, ABC employs operators that tolerate modifications over the full search space for each parameter, avoiding typical oscillations around the optimum produced by BFOA [26]. The performance of ABC algorithm has been compared with other optimization methods such as GA, DE and PSO [27,28]. The results showed that ABC can produce optimal solutions and thus is more effective than other methods in several optimization problems. Such characteristics have motivated the use of ABC to solve different sorts of engineering problems such as signal processing [29,30], flow shop scheduling [31], structural inverse analysis [32], clustering [33,34], vehicle path planning [35] and electromagnetism [36].

This paper presents an algorithm for the automatic detection of multiple circular shapes from complicated and noisy images, which does not take into consideration the conventional-Hough transform principles. The detection process is approached as a multi-modal optimization problem. The ABC algorithm searches the entire edge-map looking for circular shapes by considering a combination of three non-collinear edge points that represent candidate circles (food source locations) in the edge-only image of the scene. An objective function is used to measure the existence of a candidate circle over the edge-map. Guided by the values of such an objective function, the set of encoded candidate circles are evolved through the ABC algorithm so that the best candidate can be fitted into the most circular shape within the edge-only image. A subsequent analysis of the incorporated exhausted-source memory is then executed in order to identify potential useful local optima (other circles). The approach generates a fast sub-pixel detector that can effectively identify multiple circles in real images despite circular objects exhibiting significant occluded sections. Experimental evidence shows the effectiveness of the method for detecting circles under various conditions. A comparison with one state-of-the-art GA-based method [15], the BFOA [16] and the RHT algorithm [12] on different images has been included to demonstrate the performance of the proposed approach. Conclusions of the experimental comparison are validated through statistical tests that support the discussion suitably.

The paper is organized as follows: Section 2 provides information regarding the ABC algorithm. Section 3 depicts the implementation of the proposed circle detector. The complete multiple-circle detection procedure is presented in Section 4. Experimental outcomes after applying the proposed approach are stated in Section 5 and some relevant conclusions are discussed in Section 6.

## 2. Artificial Bee Colony (ABC) algorithm

The ABC algorithm assumes the existence of a set of operations that may resemble some features of the honey bee behavior. For instance, each solution within the search space includes a parameter set representing food source locations. The "fitness value" refers to the food source quality that is strongly linked to the food's location. The process mimics the bee's search for valuable food sources yielding an analogous process for finding the optimal solution.

*2.1 Biological bee profile*

The minimal model for a honey bee colony consists of three classes: employed bees, onlooker bees and scout bees. The employed bees will be responsible for investigating the food sources and sharing the information with recruit onlooker bees. They, in turn, will make a decision on choosing food sources by considering such information. The food source having a higher quality will have a larger chance to be selected by onlooker bees than those showing a lower quality. An employed bee, whose food source is rejected as low quality by employed and onlooker bees, will change to a scout bee to randomly search for new food sources. Therefore, the exploitation is driven by employed and onlooker bees while the exploration is maintained by scout bees. The implementation details of such bee-like operations in the ABC algorithm are described in the next sub-section.

*2.2 Description of the ABC algorithm*

Resembling other swarm based approaches, the ABC algorithm is an iterative process. It starts with a population of randomly generated solutions or food sources. The following three operations are applied until a termination criterion is met [28]:

1. Send the employed bees.
2. Select the food sources by the onlooker bees.
3. Determine the scout bees.

*2.2.1 Initializing the population*

The algorithm begins by initializing $N_p$ food sources. Each food source is a *D*-dimensional vector containing the parameter values to be optimized, which are randomly and uniformly distributed between the pre-specified lower initial parameter bound $x_j^{low}$ and the upper initial parameter bound $x_j^{high}$.

$$x_{j,i} = x_j^{low} + \text{rand}(0,1) \cdot (x_j^{high} - x_j^{low}); \tag{1}$$
$$j = 1, 2, \ldots, D; \quad i = 1, 2, \ldots, N_p.$$

with *j* and *i* being the parameter and individual indexes respectively. Hence, $x_{j,i}$ is the *j*th parameter of the *i*th individual.

*2.2.2 Send employed bees*

The number of employed bees is equal to the number of food sources. At this stage, each employed bee generates a new food source in the neighborhood of its present position as follows:

$$v_{j,i} = x_{j,i} + \phi_{j,i}(x_{j,i} - x_{j,k}); \tag{2}$$
$$k \in \{1, 2, \ldots, N_p\}; j \in \{1, 2, \ldots, D\}$$

$x_{j,i}$ is a randomly chosen *j* parameter of the *i*th individual and *k* is one of the $N_p$ food sources, satisfying the condition $i \neq k$. If a given parameter of the candidate solution $v_i$ exceeds its predetermined boundaries, that parameter should be adjusted in order to fit the appropriate range. The scale factor $\phi_{j,i}$ is a random number between $[-1,1]$. Once a new solution is generated, a fitness value representing the profitability associated with a particular solution is calculated. The fitness value for a minimization problem can be assigned to each solution $v_i$ by the following expression:

$$fit_i = \begin{cases} \dfrac{1}{1+J_i} & \text{if } J_i \geq 0 \\ 1 + abs(J_i) & \text{if } J_i < 0 \end{cases} \tag{3}$$

where $J_i$ is the objective function to be minimized. A greedy selection process is thus applied between $v_i$ and $x_i$. If the nectar- amount (fitness) of $v_i$ is better, then the solution $x_i$ is replaced by $v_i$; otherwise, $x_i$ remains.

*2.2.3 Select the food sources by the onlooker bees*

Each onlooker bee (the number of onlooker bees corresponds to the food source number) selects one of the proposed food sources, depending on their fitness value, which has been recently defined by the employed bees. The probability that a food source will be selected can be obtained from the following equation:

$$Prob_i = \dfrac{fit_i}{\sum_{i=1}^{N_p} fit_i} \tag{4}$$

where $fit_i$ is the fitness value of the food source *i*, which is related to the objective function value ($J_i$) corresponding to the food source *i*. The probability of a food source being selected by onlooker bees increases with an increase in the fitness value of the food source. After the food source is selected, onlooker bees will go to the selected food source and select a new candidate food source position inside the neighborhood of the selected food source. The new candidate food source can be expressed and calculated by (2). In case the nectar-amount, i.e., fitness of the new solution, is better than before, such position is held; otherwise, the last solution remains.

*2.2.4 Determine the scout bees*

If a food source *i* (candidate solution) cannot be further improved through a predetermined trial number known as "limit", the food source is assumed to be abandoned and the corresponding employed or onlooker bee becomes a scout. A scout bee explores the searching space with no previous information, i.e., the new solution is generated randomly as indicated by (1). In order to verify if a candidate solution has reached the predetermined "*limit*", a counter $A_i$ is assigned to each food source *i*. Such a counter is incremented consequent to a bee-operation failing to improve the food source's fitness.

*2.3 Exhausted-source memory*

Though the classic ABC algorithm eliminates the abandoned food sources, our approach considers the exhausted food sources (solutions) through avoiding their discharge by saving them into the exhausted-source memory. Such recorded solutions contain valuable information regarding global and local optima that emerged during the optimization process.

## 3. Circle detection using ABC

*3.1 Data preprocessing*

The ABC circle detector involves a pre-processing stage that requires marking the object's contour by applying a single-pixel edge detection method. For our purpose, such a task is accomplished by the classical Canny algorithm. Then, the locations of the found edge pixels are stored within the vector $P = \{p_1, p_2, \ldots, p_{E_p}\}$, $E_p$ being the total number of edge pixels in the image.

*3.2 Individual representation*

In order to construct each candidate circle $C$ (or food-source within the ABC framework), indexes $i_1$, $i_2$ and $i_3$ representing three edge points previously stored in vector $P$ must be combined. Therefore, each food-source is encoded as one circle $C = \{p_{i_1}, p_{i_2}, p_{i_3}\}$, which is characterized by three points $p_{i_1}$, $p_{i_2}$ and $p_{i_3}$ that lie on its own circumference. Such candidate circle is labeled as a potential solution for the detection problem. Considering the configuration of the edge points in Fig. 1, the circle centre $(x_0, y_0)$ and the radius $r$ of $C$ can be calculated using simple geometric equations [37].

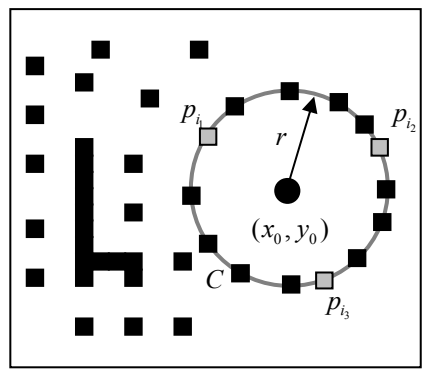

**Fig. 1.** Circle candidate (individual) built from the combination of points $p_{i_1}$, $p_{i_2}$ and $p_{i_3}$.

*3.3 Objective function*

In order to calculate the matching error produced by a candidate circle *C*, its circumference coordinates are calculated as a virtual shape that must be validated, i.e., confirm whether *C* really exists in the edge-map. Such circumference coordinates are grouped within the test set $S = \{s_1, s_2, \ldots, s_{N_s}\}$, with $N_s$ representing the number of points over which the existence of an edge point, corresponding to *C*, should be verified.

In this approach, the set *S* is generated by the midpoint circle algorithm (MCA) [38]. The MCA, which is considered to be the quickest method providing a sub-pixel precision [39,40], calculates the required points for digitally drawing a circle. Considering that the function $f_{Circle}(x,y) = x^2 + y^2 - r^2$ defines a circular primitive, MCA introduces an error *e* as a measurement for the deviation of the halfway pixel position (sub-pixel distance) characterized by $e = f_{Circle}(x,y)$, with *e* being zero for locations lying on the circumference, positive for those outside and negative for those occupying the interior. The minimum error (i.e., the error shown by the pixel lying closer to the ideal circumference) is used to decide which pixel should be set next as the best circle boundary approximation. On the other hand, the computation time of MCA is reduced by considering the symmetry among circles. Circle sections lying at adjacent octants within one quadrant, are symmetric with respect to the 45° line dividing two octants. Taking advantage of such symmetry property, MCA generates all pixel positions around a circle by calculating only the first octant. Therefore, other octants are inferred from the first one by using simple symmetry relationships. For more details, see [41].

The objective function *J(C)* represents the matching error produced between the pixels *S* (calculated by MCA) of the circle candidate *C* and the pixels that actually exist in the edge image, yielding:

$$J(C) = 1 - \frac{\sum_{v=1}^{Ns} E(s_v)}{Ns} \qquad (5)$$

where $E(s_v)$ is a function that verifies the pixel existence in the $s_v$ location ($s_v = (x_v, y_v)$), with $s_v \in S$ and $N_s$ being the number of pixels lying on the perimeter corresponding to *C* currently under testing. Hence, the function $E(s_v)$ is defined as:

$$E(s_v) = \begin{cases} 1 & \text{if the pixel } (x_v, y_v) \text{ is an edge point} \\ 0 & \text{otherwise} \end{cases} \qquad (6)$$

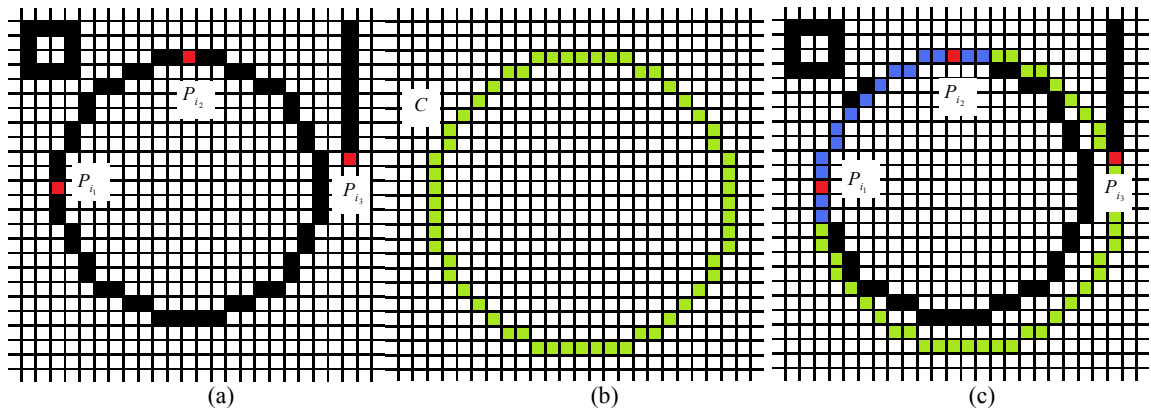

**Fig. 2.** Procedure to evaluate the objective function *J(C)*: (a) the original edge-map, (b) the virtual shape generated by MCA considering $C = \{p_{i_1}, p_{i_2}, p_{i_3}\}$. (c) shows the comparison operation between the original edge-map shown in (a) and the virtual shape presented by (b).

A value near to zero of *J(C)* implies a better response from the "circularity" operator and represents a better nectar- amount within the ABC framework. Fig. 2 shows the procedure to evaluate the objective function *J(C)*. First, three edge points (as they are exhibited by Fig. 2a) encode a candidate circle *C*. Then, by using the MCA, a circular virtual shape is built (see Fig. 2b). Such virtual shape, which is characterized by the vector *S*, groups a determined number $N_s$ of pixel coordinates that are shown by Fig. 2b as 56. Finally, the virtual shape is compared with the original edge-map, point by point, in order to find matches between virtual and edge pixels. Fig. 2c shows the comparison operation between the original edge-map (Fig. 2a) and the virtual shape (Fig. 2b). As a result, only eighteen edge pixels are common to both images yielding: $\sum_{v=1}^{Ns} E(x_v, y_v) = 18$ and therefore, $J(C) \approx 0.67$.

*3.3. ABC Implementation*

The implementation of the proposed algorithm can be summarized by the following steps:

**Step 1:** Apply the Canny filter to the original image and store edge pixels within vector *P*.

**Step 2:** Initialize required parameters of the ABC algorithm. Set the colony's size, the abandonment *limit* and the maximum number of cycles.

**Step 3:** Initialize $N_C$ circle candidates $C_b$ (original food sources) with $b \in (1,\ldots,N_C)$, and clear all counters $A_b$.

**Step 4:** Obtain the matching fitness (food source quality) for each circle candidate $C_b$ using (5).

**Step 5:** Repeat steps 6 to 10 until a termination criterion is met.

**Step 6:** Modify the circle candidates as stated by (2) and evaluate its matching fitness (send employed bees onto food sources). Likewise, update all counters $A_b$.

**Step 7** Calculate the probability value $Prob_b$ for each circle candidate $C_b$. Such probability value will be used as a preference index by onlooker bees (4).

**Step 8:** Generate new circles candidates (using the (2)) from current candidates according to their probability $Prob_b$ (Send onlooker bees to their selected food source). Likewise, update counters $A_b$.

**Step 9:** Obtain the matching fitness for each circle candidate $C_b$ and calculate the best circle candidate (solution).

**Step 10:** Stop modifying the circle candidate $C_b$ (food source) whose counter $A_b$ has reached its counter "*limit*" and save it as a possible solution (global or local optimum) in the exhausted-source memory. Clear $A_b$ and generate a new circle candidate according to (1).

**Step 11:** Analyze solutions previously stored in the exhausted-source memory (see Section 5). The memory holds solutions (any other potential circular shape in the image) generated through the evolution of the optimization algorithm.

In ABC algorithm, the steps 6 to 10 are repeated until a termination criterion is met. Typically, two stop criteria have been employed for meta-heuristic algorithms: either an upper limit of the fitness function is reached or an upper limit of the number of generations is attained [42]. The first criterion requires an extensive knowledge of the problem and its solutions [43]. On the contrary, by considering the stop criterion based on the number of generations, feasible solutions may be found by exploring the search space through several iterations. For our purpose, the number of iterations as stop criterion is employed in order to allow the multi-circle detection. Hence, if a solution representing a valid circle appears at early stages, it would be stored in the exhausted-source memory and the algorithm continues detecting other feasible solutions until depleting the iteration number. Therefore, the main issue is to define a fair iteration number, which should be big enough to allow finding circles at the image and small enough to avoid an exaggerated computational cost. For this study, such a number was experimentally defined as 300.

**4. The multiple-circle detection procedure**

The original ABC algorithm considers the so-called abandonment limit, which aims to stop the local exploration for a candidate solution after a trial number is reached. All "stuck solutions", i.e., those that do not improve further during the optimization cycle are supposed to be discarded and replaced by other randomly generated solutions. However, this paper proposes the use of an "exhausted-source memory" to store information regarding local optima that represent possible solutions for the multi-circle detection problem.

Several heuristic methods have been employed for detecting multiple circles as an alternative to classical Hough transform-based techniques [15, 16]. Such strategies imply that only one circle can be marked per optimization cycle, forcing a multiple execution of the algorithm in order to achieve multiple-circle detection. The surface representing $J(C)$ holds a multimodal nature, which contains several global and local optima that are related to potential circular shapes in the edge-map. This paper aims to solve the objective function $J(C)$ using only one optimization procedure by assuming the multi-detection problem as a multimodal optimization issue.

The multi-detection problem can be summarized as follows: guided by the values of a matching function, the set of encoded circle candidates are evolved through the ABC algorithm and the best circle candidate (global optimum) is considered to be the first detected circle over the edge-only image. Then, an analysis of the incorporated exhausted-source memory is executed in order to identify other local optima (other circles). The analysis includes two operations: arraignment and extraction. In the arraignment, food sources that are held by the memory are organized in descending order according to their $J(C)$. Once the exhausted-source memory has been arranged, the goal is to extract circles considered to be different (local optima) from it. Such discrimination is accomplished by comparing all elements in the arranged memory.

Several local optima (i.e., circles slightly shifted or holding small deviations) can represent the same circle. Therefore, a distinctiveness factor $E_{s_{di}}$ is required to measure the mismatch between two given circles (food-sources) as follows:

$$E_{s_{di}} = \sqrt{(x_A - x_B)^2 + (y_A - y_B)^2 + (r_A - r_B)^2} \tag{7}$$

where $(x_A, y_A)$ and $r_A$ are the coordinates of the centre and radius of the circle $C_A$ respectively, while $(x_B, y_B)$ and $r_B$ are the corresponding parameters of the circle $C_B$. In order to decide whether two circles must be considered different, a threshold value $E_{s_{TH}}$ is defined as follows:

$$E_{S_{TH}} = \alpha \sqrt{(cols - 1)^2 + (rows - 1)^2 + (r_{\max} - r_{\min})^2} \tag{8}$$

where *rows* and *cols* refer to the number of rows and columns in the image respectively. $r_{\max}$ and $r_{\min}$ are the maximum and minimum radii for representing feasible candidate circles, while $\alpha$ is a sensitivity factor affecting the discrimination between circles. A high value of $\alpha$ allows circles to be significantly different and still be considered as the same shape while a low value would imply that two circles with slight differences in radii or positions could be considered as different instances. The $E_{s_{TH}}$ value calculated by (8) allows discriminating circles with no consideration about the image size.

In order to find "sufficiently different" circles, the elements of the arranged exhausted-source memory must be contrasted. Each element is compared with the others by using (7). A circle is considered different enough if its distinctiveness factor $E_{s_{di}}$ (found in the comparison) surpasses the threshold $E_{s_{TH}}$.

The multiple-circle detection procedure can thus be described as follows:

**Step 1:** The best solution found by the ABC algorithm and all candidate circles held by the exhausted-source memory are organized in a decreasing order of their matching fitness, yielding a new vector $M_C = \{C_1, \ldots, C_{Ne}\}$, with *Ne* being the size of the exhausted-source memory plus one.

**Step 2:** The candidate circle $C_1$ showing the highest matching fitness is identified as the first circle shape $CS_1$ as it is stored within a vector $Ac$ of actual circles.

**Step 3:** The distinctiveness factor $E_{s_{di}}$ for the candidate circle $C_m$ (element *m* in $M_C$) is compared with every element in $Ac$. If $E_{s_{di}} > E_{S_{TH}}$ is true for each pair of solutions (those present in $Ac$ and in the candidate circle $C_m$), then $C_m$ is considered as a new circle shape $CS$ and is added to the vector $Ac$. Otherwise, the next circle candidate $C_{m+1}$ is evaluated and $C_m$ is discarded.

**Step 4:** Step 3 is repeated until all $Ne$ candidate circles in $M_C$ have been analyzed.

Summarizing the overall procedure, Fig. 3 shows the outcome of the ABC-based circular detector. The input image (Fig. 3a) has a resolution of 256 x 256 pixels and shows two circles and two ellipses with a different circularity factor. Fig. 3b presents the detected circles with a red overlay. Fig. 3c shows the candidate circles held by the exhausted-source memory after the optimization process. Fig. 3d presents the resulting image after the previously described discrimination procedure has been applied.

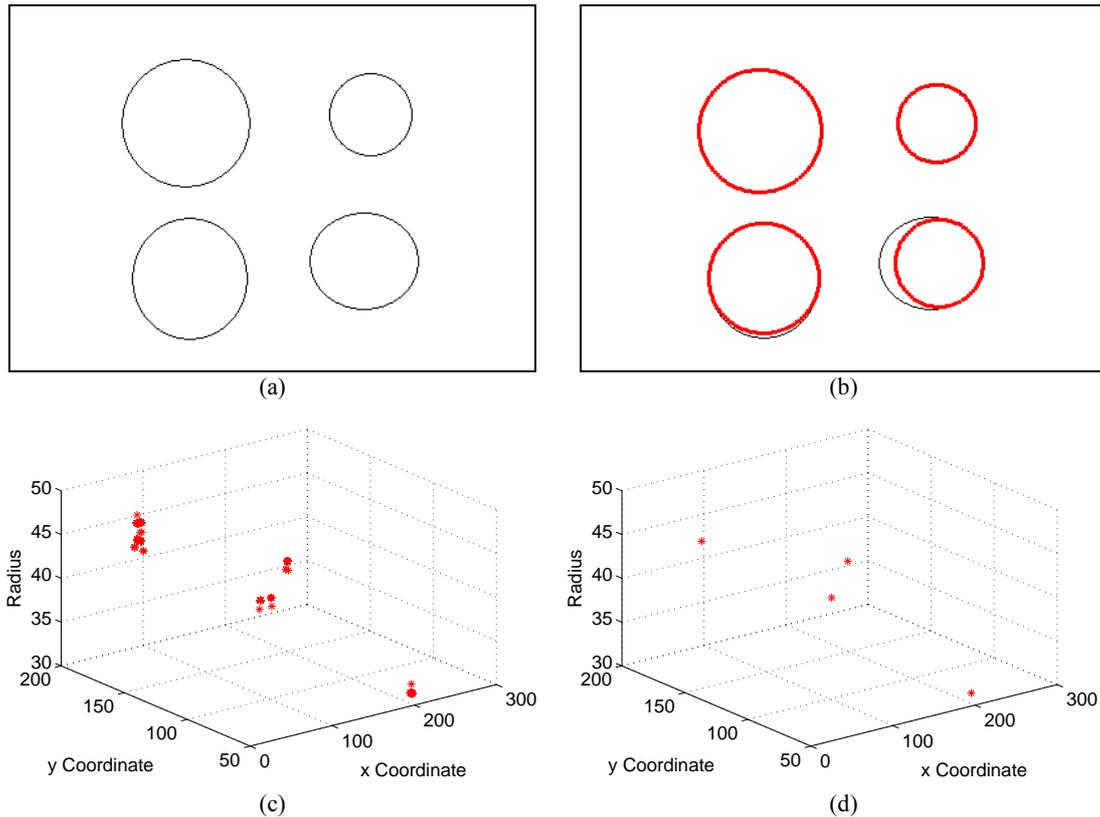

**Fig. 3.** ABC-based circular detector performance: (a) the original image and (b) all detected circles as an overlay (c) Candidate circles held by the exhausted-source memory after optimization process and (d) remaining circles after the discrimination process.

## 5. Experimental Results

Experimental tests have been developed in order to evaluate the performance of the circle detector. The experiments address the following tasks:

(1) Circle localization,
(2) Shape discrimination,
(3) Circular approximation: occluded circles and arc detection.

Table 1 presents the parameters for the ABC algorithm in this study. They have been retained for all test images after being experimentally defined.

| Colony size | Abandonment limit | Number of cycles | $\alpha$ | *limit* |
|---|---|---|---|---|
| 20 | 100 | 300 | 0.05 | 30 |

**Table 1.** ABC detector parameters

All the experiments have been executed over a Pentium IV 2.5 GHz computer under C language programming. All the images are pre-processed by the standard Canny edge detector using the image-processing toolbox for MATLAB R2008a.

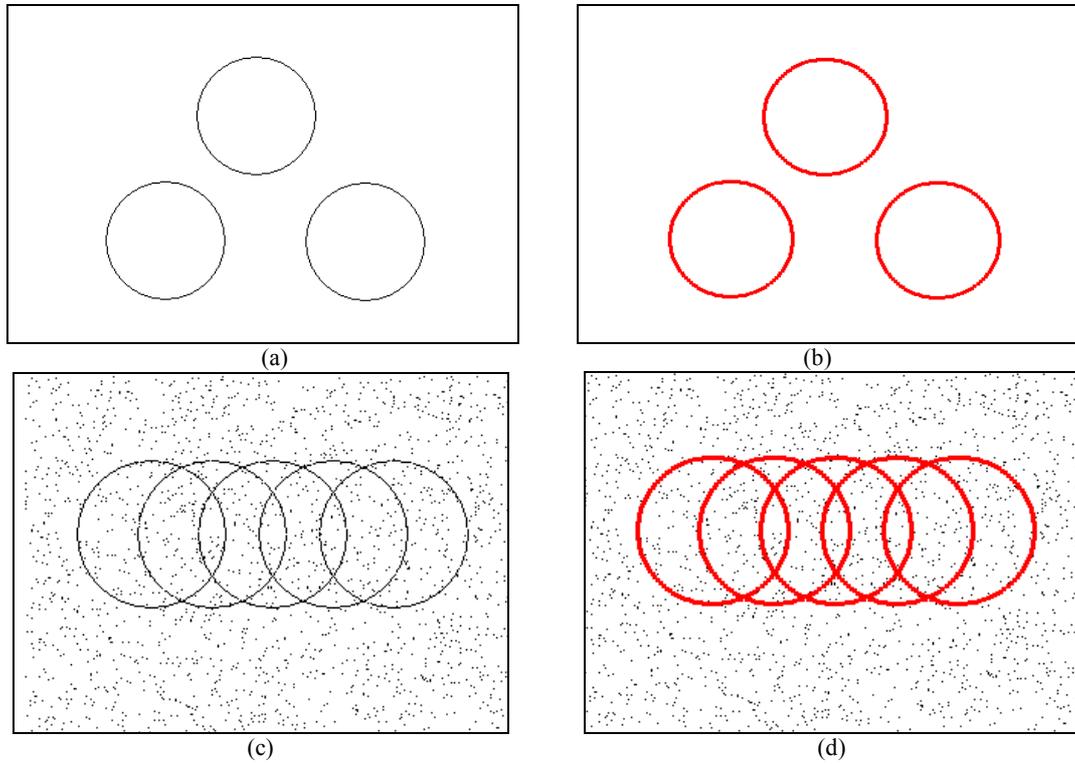

**Fig. 4.** Circle localization over synthetic images. The image (a) shows the original image while (b) presents the detected circles as an overlay. The image in (c) shows a second image with salt & pepper noise and (d) shows detected circles as a red overlay.

*5.1 Circle localization*

5.1.1. Synthetic images

The experimental setup includes the use of several synthetic images of 320x240 pixels. All images contain varying amounts of circular shapes and some have also been contaminated by added noise so as to increase the complexity of the localization task. The algorithm is executed over 100 times for each test image, successfully identifying and marking all circles in the image. The detection has proved to be robust to translation and scaling, requiring less than 1s. Fig. 4 shows the outcome after applying the algorithm over two images taken from the experimental set.

5.1.2. Natural images

This experiment tests the circle detection on real-life images. Twenty five test images of 640x480 pixels have been captured using a digital camera under an 8-bit color format. Each natural scene includes circular shapes that have been pre-processed through the Canny edge detection algorithm before being fed to the ABC procedure. Fig. 5 shows a multiple-circle detection over a natural image.

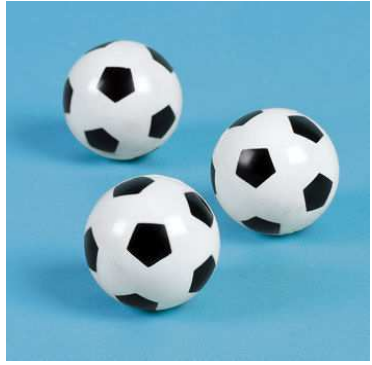 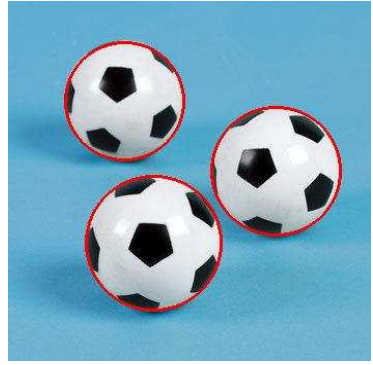

(a) (b)

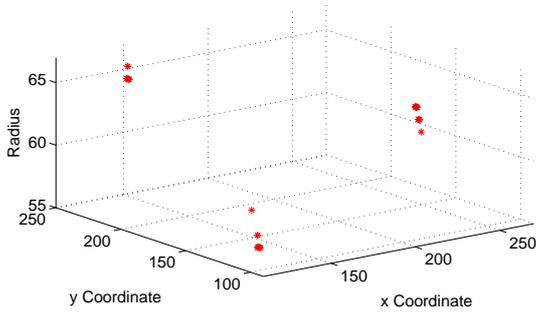 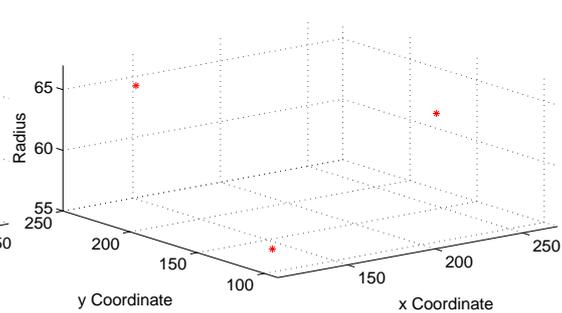

(c) (d)

**Fig. 5.** Circle detection algorithm over natural images: (a) the original image (b) the detected circles as a red overlay (c) candidate circles lying at the exhausted-source memory after the optimization and (d) detected circles after finishing the discrimination process.

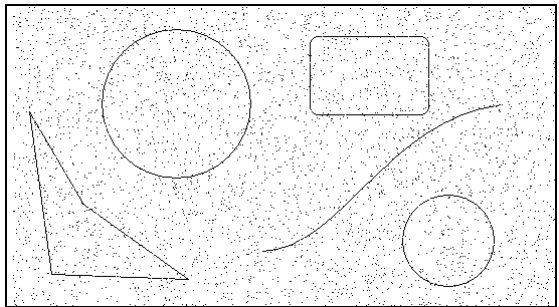 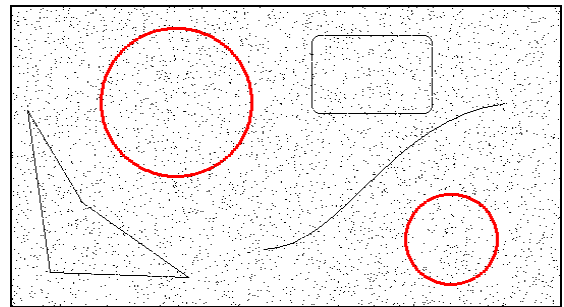

(a) (b)

**Fig. 6.** Shape discrimination over synthetic images: (a) the original image contaminated by salt & pepper noise (b) detected circles as an overlay.

*5.2. Shape discrimination tests*

This section discusses the detector's ability to differentiate circular patterns over any other shape, which might be present in the image. Fig. 6 shows five different synthetic shapes within an image of 540x300 pixels that has been contaminated by salt & pepper noise. Fig. 7 repeats the experiment over real-life images.

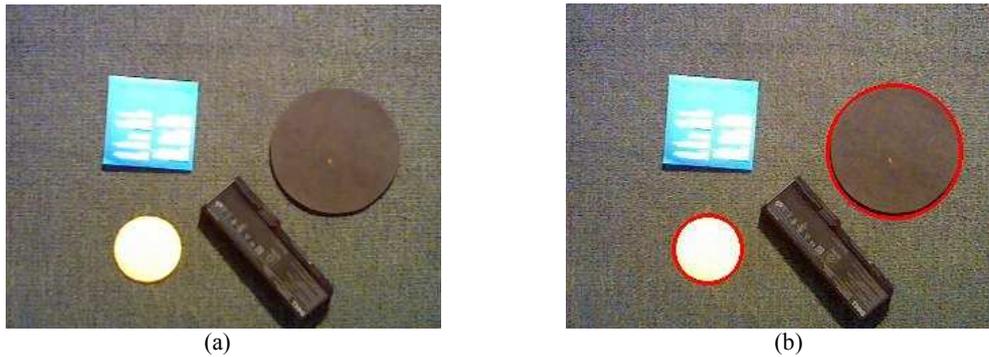

(a)                                        (b)
**Fig. 7.** Shape discrimination in real-life images: (a) original image and (b) the detected circle as an overlay.

*5.3 Circular approximation: occluded circles and arc detection.*

The ABC detector algorithm is able to detect occluded or imperfect circles as well as partially defined shapes such as arc segments. The relevance of such functionality comes from the fact that imperfect circles are commonly found in typical computer vision applications. Since circle detection has been considered as an optimization problem, the ABC algorithm allows finding circles that may approach a given shape according to fitness values for each candidate. Fig. 8a shows some examples of circular approximation. Likewise, the proposed algorithm is able to find circle parameters that better approach an arc or an occluded circle. Fig. 8b and 8c show some examples of this functionality. A small value for *J(C)*, i.e., near zero, refers to a circle while a slightly bigger value accounts for an arc or an occluded circular shape. Such a fact does not represent any trouble as circles can be shown following the obtained *J(C)* values.

*5.4 Performance evaluation*

In order to enhance the algorithm analysis, the ABC proposed algorithm is compared with the BFAOA and the GA circle detectors over a set of common images.

The GA algorithm follows the proposal of Ayala-Ramirez et al., which considers the population size as 70, the crossover probability as 0.55, the mutation probability as 0.10 and the number of elite individuals as 2. The roulette wheel selection and the 1-point crossover operator are both applied. The parameter setup and the fitness function follow the configuration suggested in [15]. The BFAOA algorithm follows the implementation from [16] considering the experimental parameters as: $S=50$, $N_c=100$, $N_s=4$, $N_{ed}=1$, $P_{ed}=0.25$, $d_{attract}=0.1$, $w_{attract}=0.2$, $w_{repellant}=10$ $h_{repellant}=0.1$, $\lambda=400$ and $\psi=6$. Such values are found to represent the best configuration set according to [16].

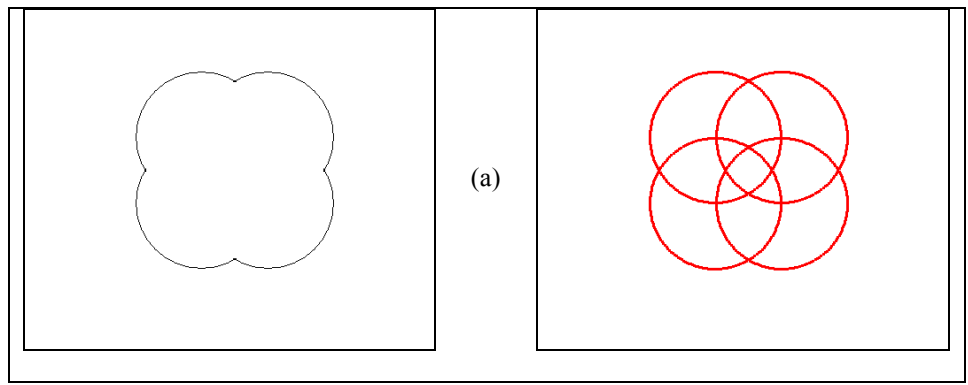

(a)

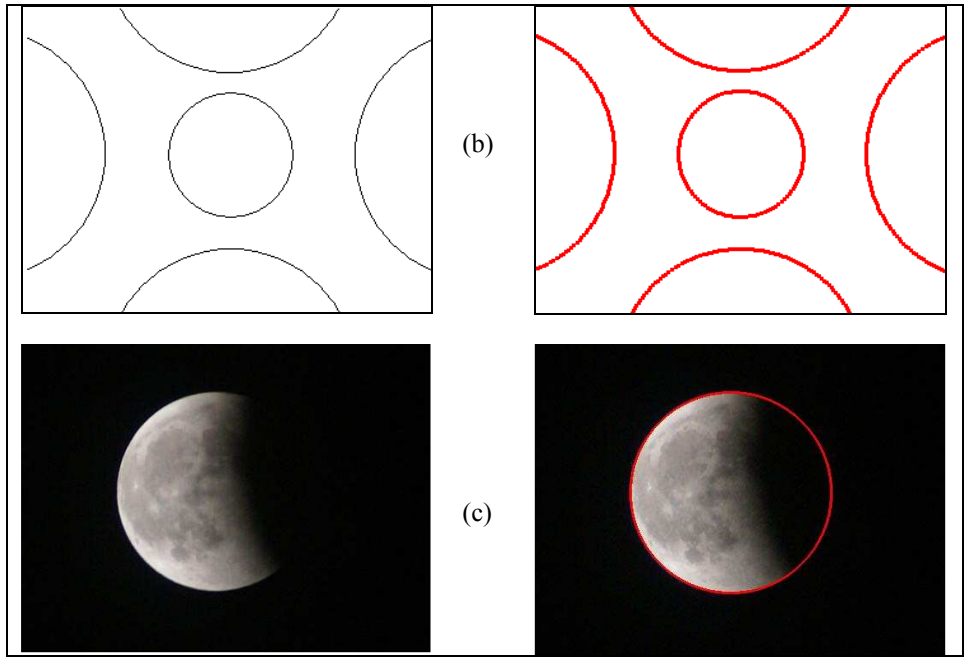

**Fig. 8.** ABC approximating circular shapes and arc sections.

Images rarely contain perfectly-shaped circles. Therefore, aiming for a test on accuracy for a single-circle, the detection is challenged by a ground-truth circle, which is determined manually from the original edge-map. The parameters $(x_{true}, y_{true}, r_{true})$ of the reference circle are computed considering the best matching circle that has been previously defined over the edge-map. If the centre and the radius of the detected circle are defined as $(x_D, y_D)$ and $r_D$, then an error score (Es) can be chosen as follows:

$$\text{Es} = \eta \cdot (|x_{true} - x_D| + |y_{true} - y_D|) + \mu \cdot |r_{true} - r_D| \qquad (9)$$

The central point difference $(|x_{true} - x_D| + |y_{true} - y_D|)$ represents the centre shift for the detected circle as it is compared with a benchmark circle. The radio mismatch $(|r_{true} - r_D|)$ accounts for the difference between their radii. $\eta$ and $\mu$ represent two weighting parameters, which are to be applied separately to the central point difference and to the radius mismatch for the final error Es. In this study, they are chosen as $\eta = 0.05$ and $\mu = 0.1$. This particular choice ensures that the radii difference would be strongly weighted in comparison to the difference in the central circular positions of manually detected and machine-detected circles. In order to use an error metric for multiple-circle detection, the averaged Es produced from each circle in the image is considered. Such criterion, defined as the multiple error (ME), is calculated as follows:

$$\text{ME} = \left(\frac{1}{NC}\right) \cdot \sum_{R=1}^{NC} \text{Es}_R \qquad (10)$$

where *NC* represents the number of circles actually present the image. In case the ME is less than 1, the algorithm is considered successful; otherwise it is said to have failed in the detection of the circle set. Notice that for $\eta = 0.05$ and $\mu = 0.1$, it yields ME<1, which accounts for a maximal tolerated average difference on radius length of 10 pixels, whereas the maximum average mismatch for the centre location can be up to 20 pixels. In general, the success rate (SR) can thus be defined as the percentage of achieving success after a certain number of trials.

|     | **Original image** | **Best case** | **Worst case** |
| --- | --- | --- | --- |
| (a) | | | |
| (b) | | | |
| (c) | | | |
| (d) | | | |
| (e) | | | |
| (f) | | | |

**Fig. 9.** Test images and their detected circles using the ABC detector. The results show the best and the worst case obtained throughout 35 runs.

Fig. 9 shows six images that have been used to compare the performance of the GA-based algorithm [15], the BFOA method [16] and the proposed approach. The performance is analyzed by considering 35 different executions for each algorithm over six images. Table 2 presents the averaged execution time, the success rate (SR) in percentage and the averaged multiple error (ME). The best entries are bold-cased in Table 2. Closer inspection reveals that the proposed method is able to achieve the highest success rate with the smallest error, and still requires less computational time for most cases. Fig. 9 also exhibits the resulting images after applying the ABC detector. Such results present the best and the worst cases obtained throughout 35 runs.

A non-parametric statistical significance proof called Wilcoxon's rank sum test for independent samples [44-46] has been conducted at 5% significance level on the multiple error (ME) data of Table 2. Table 3 reports the *p*-values produced by Wilcoxon's test for the pair-wise comparison of multiple error (ME) between two groups. One group corresponds to ABC versus GA and the other corresponds to ABC versus BFOA, one at a time. As a null hypothesis, it is assumed that there is no significant difference between the mean values of the two groups. The alternative hypothesis considers a significant difference between the mean values of both groups. All *p*-values reported in the table are less than 0.05 (5% significance level), which is a strong evidence against the null hypothesis, indicating that the best ABC mean values for the performance are statistically significant and have not occurred by chance.

| Image | Averaged execution time ± Standard deviation (s) | | | Success rate (SR) (%) | | | Averaged ME ± Standard deviation | | |
|---|---|---|---|---|---|---|---|---|---|
| | GA | BFOA | ABC | GA | BFOA | ABC | GA | BFOA | ABC |
| (a) | 2.23±(0.41) | 1.71±(0.51) | **0.21±(0.22)** | 94 | **100** | **100** | 0.41±(0.044) | 0.33±(0.052) | **0.22±(0.033)** |
| (b) | 3.15±(0.39) | 2.80±(0.65) | **0.36±(0.24)** | 81 | 95 | **98** | 0.51±(0.038) | 0.37±(0.032) | **0.26±(0.041)** |
| (c) | 4.21±(0.11) | 3.18±(0.36) | **0.20±(0.19)** | 79 | 91 | **100** | 0.48±(0.029) | 0.41±(0.051) | **0.15±(0.036)** |
| (d) | 5.11±(0.43) | 3.45±(0.52) | **1.10±(0.24)** | 93 | **100** | **100** | 0.45±(0.051) | 0.41±(0.029) | **0.25±(0.037)** |
| (e) | 6.33±(0.34) | 4.11±(0.14) | **1.61±(0.17)** | 87 | 94 | **100** | 0.81±(0.042) | 0.77±(0.051) | **0.37±(0.055)** |
| (f) | 7.62±(0.97) | 5.36±(0.17) | **1.95±(0.41)** | 88 | 90 | **98** | 0.92±(0.075) | 0.88±(0.081) | **0.41±(0.066)** |

**Table 2.** The averaged execution-time, success rate and the averaged multiple error for the GA-based algorithm, the BFOA method and the proposed ABC algorithm, considering the six test images shown in Fig. 9

| Image | *p*-Value | |
|---|---|---|
| | ABC vs. GA | ABC vs. BFOA |
| (a) | 1.8061e-004 | 1.8288e-004 |
| (b) | 1.7454e-004 | 1.9011e-004 |
| (c) | 1.7981e-004 | 1.8922e-004 |
| (d) | 1.7788e-004 | 1.8698e-004 |
| (e) | 1.6989e-004 | 1.9124e-004 |
| (f) | 1.7012e-004 | 1.9081e-004 |

**Table 3.** *p*-values from Wilcoxon's test, comparing ABC with GA and BFOA over the ME from Table 2.

Fig. 10 demonstrates the relative performance of ABC in comparison with the RHT algorithm following the proposal in [12]. All images belonging to the test are complicated and contain different noise conditions. The performance analysis is achieved by considering 35 different executions for each algorithm over the three images. The results, exhibited in Fig. 10, present the median-run solution (when the runs were ranked according to their final ME value) obtained throughout the 35 runs. On the other hand, Table 4 reports the corresponding averaged execution time, success rate (in %), and average multiple error (using (10)) for ABC and RHT algorithms over the set of images. Table 4 shows a decrease in performance of the RHT algorithm as noise conditions change. Yet the ABC algorithm holds its performance under the same circumstances.

| Image | Average time ± Standard deviation (s) | | Success rate (SR) (%) | | Average ME ± Standard deviation | |
|---|---|---|---|---|---|---|
| | RHT | ABC | RHT | ABC | RHT | ABC |
| (I) | 7.82±(0.34) | **0.20±(0.31)** | 100 | 100 | 0.19±(0.041) | **0.20±(0.021)** |
| (II) | 8.65±(0.48) | **0.23±(0.28)** | 70 | 100 | 0.47±(0.037) | **0.18±(0.035)** |
| (III) | 10.65±(0.48) | **0.22±(0.21)** | 18 | 100 | 1.21±(0.033) | **0.23±(0.028)** |

**Table 4.** Average time, success rate and averaged error for ABC and HT, considering three test images

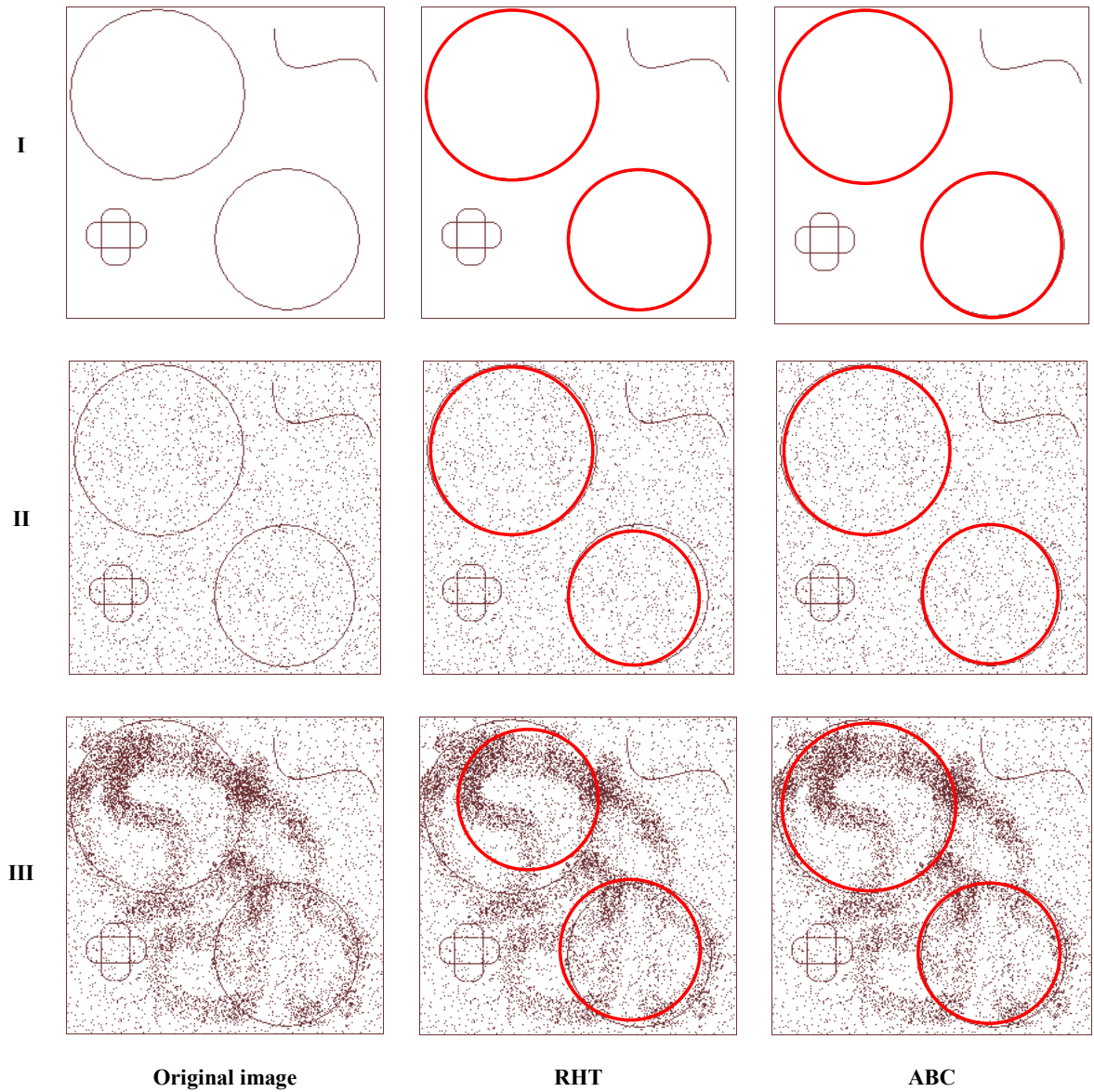

**Original image**  **RHT**  **ABC**

**Fig. 10.** Relative performance of RHT and ABC

## 6. Conclusions

This paper has presented an algorithm for the automatic detection of multiple circular shapes from complicated and noisy images without considering the conventional Hough Transform principles. The detection process is considered to be similar to a multi-modal optimization problem. In contrast to other heuristic methods that employ an iterative procedure, the proposed ABC method is able to detect single or multiple circles over a digital image by running only one optimization cycle. The ABC algorithm searches the entire edge-map for circular shapes by using a combination of three non-collinear edge points as candidate circles (food positions) in the edge-only image. A matching function (objective function) is used to measure the existence of a candidate circle over the edge-map. Guided by the values of this matching function, the set of encoded candidate circles is evolved using the ABC algorithm so that the best candidate can fit into an actual circle. A novel contribution is related to the exhausted-source memory that has been designed to hold "stuck" solutions which, in turn, represent feasible solutions for the multi-circle detection problem. A post-analysis on the exhausted-source memory should indeed detect other local minima, i.e., other potential circular shapes. The overall approach generates a fast sub-pixel detector, which can effectively identify multiple circles in real images despite circular objects exhibiting a significant occluded portion.

Classical Hough Transform methods for circle detection use three edge points to cast a vote for the potential circular shape in the parameter space. However, they would require a huge amount of memory and longer computational times to obtain a sub-pixel resolution. Moreover, HT-based methods rarely find a precise parameter set for a circle in the image [47]. In our approach, the detected circles hold a sub-pixel accuracy inherited directly from the circle equation and the MCA method.

In order to test the circle detection performance, both speed and accuracy have been compared. Score functions are defined by (9)-(10) in order to measure accuracy and effectively evaluate the mismatch between manually-detected and machine-detected circles. We have demonstrated that the ABC method outperforms both the GA (as described in [15]) and the BFOA (as described in [16]) within a statistically significant framework (Wilcoxon test). In contrast to the ABC method, the RHT algorithm [12] shows a decrease in performance under noisy conditions. Yet the ABC algorithm holds its performance under the same circumstances.

Finally, Table 2 indicates that the ABC method can yield better results on complicated and noisy images compared with the GA and the BFOA methods. However, the aim of this study is not to beat all the circle detector methods proposed earlier, but to show that the ABC algorithm can effectively serve as an attractive method to successfully extract multiple circular shapes.